\documentclass[11pt]{article}

\usepackage{acl}

\usepackage{times}
\usepackage{latexsym}
\usepackage[T1]{fontenc}
\usepackage[utf8]{inputenc}
\usepackage{microtype}
\usepackage{graphicx}
\usepackage{overpic}
\usepackage{booktabs}
\usepackage{amsmath}
\usepackage{amssymb}
\usepackage{array}
\usepackage{multirow}

\newcommand{\pp}{percentage points}

\title{A Probe Shift Is Not a Fairness Fix: The Limits of Representation Steering in Speech Models}

\author{Nicolas Bourrel \qquad Abderrahmane Issam \qquad {\bf Gerasimos Spanakis} \\
        Department of Advanced Computing Sciences \\ 
        Maastricht University \\ 
        \small{\texttt{\{n.bourrel@student., abderrahmane.issam@, jerry.spanakis@\}maastrichtuniversity.nl}}}
\begin{document}
\maketitle

\begin{abstract}
Automatic speech recognition (ASR) systems exhibit unequal error rates across speaker groups, motivating interventions on their internal representations.
We ask whether speaker-linked attributes that are linearly readable from pretrained ASR encoders yield useful directions for reducing group word-error-rate (WER) gaps.
Across Whisper-medium, HuBERT-large, and Wav2Vec2-large on Common Voice and the Speech Accent Archive, we probe every encoder layer for metadata-derived sex/gender, age, and native/accent labels; construct centroid and probe-derived directions; inject them at selected layers; and compare downstream probe trajectories with matched WER changes.
Sex labels are highly decodable (best macro-F1 0.924--0.941), native/accent labels are also above chance (0.544--0.696), and age is weaker (0.354--0.397).
Of 22 post-selected reruns, nine have 95\% paired-bootstrap intervals entirely below zero, yet every absolute source-group WER reduction is below 0.7 percentage points.
Conversely, a local target-class probe rate can rise from 8.09\% to 99.87\% while WER worsens.
Linear readability is therefore neither evidence of causal use nor a reliable mitigation method.
Our results motivate evaluating speech-bias interventions jointly at representation, propagation, and task levels.
\end{abstract}

\section{Introduction}

Automatic speech recognition (ASR) can perform unevenly across accents, genders, languages, ages, and other speaker-linked groups \cite{allison2020, feng2021quantifyingbiasautomaticspeech, jahan2025}.
Such disparities can reduce access to voice interfaces and compound exclusion in education, employment, public services, and assistive technology.
Prior audits document group-level error gaps and show that speech encoders retain information associated with demographic metadata \citep{attanasio-etal-2024-twists,lin24b_interspeech, herron-etal-2026-identifying, herron2026}.
Together, these findings suggest an appealing intervention: if a disadvantaged group's attribute is visible inside an ASR encoder, move its hidden representation toward a lower-error comparison group.

That proposal combines three different claims.
A probe can show that an attribute is \emph{readable}; an intervention can make later probes read a state as more target-like; and the recogniser can produce fewer word errors.
None logically entails the next.
Probing work has long cautioned that decodability does not establish how a model uses a feature \citep{belinkov-2022-probing,kumar2022}.
For fairness interventions the distinction matters especially: a nearly complete demographic-probe shift may leave recognition unchanged, damage linguistic content, or require knowing a sensitive label at inference time.

We test this chain in three frozen ASR systems with different objectives and decoders: Whisper \cite{pmlr-v202-radford23a}, HuBERT large \cite{hsu2021}, and Wav2Vec2 \cite{Baevski2020}.
On Mozilla Common Voice (MCV) \cite{ardila-etal-2020-common} and the Speech Accent Archive (SAA) \cite{weinberger2015speech}, we first measure layer-wise linear access to metadata-derived sex/gender, age, and native/accent labels using linear probes. Then we apply activation steering from higher to lower Word Error Rate (WER) classes using contrastive steering \cite{marks2024the, rimsky-etal-2024-steering} with the centroid difference or using the linear probe direction \cite{hedstrom2025to, chen2026adaptive}. 
Finally, we evaluate the intervention at two downstream levels: whether its target-like shift propagates through later layers, and whether the steered model's generation achieves a lower WER.

This paper makes three contributions.
First, it provides a common layer-wise audit across three 24-block ASR encoders and two speech corpora, keeping the metadata, representation, and task measurements explicit.
Second, it connects descriptive probing to a controlled representation intervention without retraining the ASR model.
Third, it reports the complete set of 22 selected follow-up reruns, including null and adverse cases, and isolates a failure mode in which a probe shift is large but ASR does not improve.

The central result is cautionary.
Speaker-linked labels are often strongly readable, and fixed directions can produce dramatic target-like shifts, but WER effects are small and unstable across layers.
Nine selected reruns have paired-bootstrap intervals below zero, yet none reduces source-group WER by one point.
One Wav2Vec2 setting makes 99.87\% of source clips target-like to the next-layer probe while WER worsens.
We therefore present the method as a stress test for representation-based mitigation, not as a fairness fix.

\section{Related Work}

\subsection{Speech representations and probes}

Wav2Vec2 learns masked speech representations through contrastive prediction \citep{Baevski2020}; HuBERT predicts clustered hidden units under masking \citep{hsu2021}; and Whisper uses large-scale weak supervision in an encoder--decoder recogniser \citep{pmlr-v202-radford23a}.
Their objectives and decoding paths differ, but all can retain information beyond lexical content.
Wav2Vec2 features transfer to speaker verification and language identification \citep{fan2021}, and layer-wise analyses show that acoustic and linguistic content is distributed non-monotonically through speech encoders \citep{belinkov2017,pasad2021}.

Linear probes offer a deliberately simple test of whether a label is accessible from a hidden state \citep{alain2017understanding,belinkov-2022-probing}.
Speech probes recover phonetic, speaker, environment, emotion, and other information \citep{raymondaud2024probinginformationencodedneuralbased}.
However, a successful probe can use a correlate that the base model ignores, and classifier performance depends on representation, probe capacity, controls, and data construction.
We therefore interpret macro-F1 as linear accessibility, not a causal statement about ASR behaviour.

\subsection{Fairness and representation intervention}

Speech fairness studies measure error disparities and the encoding or propagation of social information.
For example, multilingual ASR systems exhibit gender-related performance gaps whose relationship to internal separability varies by language and model \citep{attanasio-etal-2024-twists}; self-supervised speech encoders can also encode social biases affected by architecture and compression \citep{lin24b_interspeech}.
These results motivate internal analysis but do not imply that demographic separability is the cause of an observed WER gap.

Nearby intervention work has stronger machinery than the fixed vectors studied here.
Speaker-adaptation systems learn or retrieve speaker information to improve ASR \citep{sari2020}.
Privacy-oriented models adversarially conceal attributes while preserving speaker verification \citep{noe2021}.
Other work identifies speaker subspaces and removes them to reduce speaker identifiability while retaining phonetic information \citep{liu2023}.
Our intervention is intentionally simpler: no model parameters are updated, and a single metadata-derived vector is added at one layer.
This makes it a transparent baseline for asking whether readability, movement, and task benefit actually align.

\section{Experimental Design}

\subsection{Data and operational labels}

MCV is a crowd-sourced multilingual speech corpus with speaker-contributed recordings and metadata \citep{ardila-etal-2020-common}.
We use English clips and construct task-specific subsets for sex/gender, age, and native/accent labels.
Each task begins with 100,000 training identifiers and 1,500 test identifiers before task-label filtering.
The five-class age summary retains 96,223 labelled training clips and 1,237 labelled test clips; sex/gender and native/accent each retain all 1,500 test clips.
The generated MCV subsets have no shared \texttt{client\_id} values between train and test.

SAA contains speakers reading a shared English passage and provides native-language and demographic metadata \citep{weinberger2015speech}.
We use a deterministic 80/20 metadata split, yielding 1,710 training and 428 test items after the available-audio and label filters.
The split code operates on filename stems rather than enforcing an explicit grouped-speaker split; the archive metadata nevertheless has one \texttt{speakerid} per row.
SAA is valuable for accent comparisons because lexical content is fixed, but several native-language and older-age groups are small.

Table~\ref{tab:data} summarises the evaluation sets.
The labels are operational groupings from corpus metadata, not complete measurements of identity, biology, accent, or linguistic background.
The source-to-target notation used below only encodes an observed error-rate contrast.

\begin{table*}[t]
\centering
\small
\setlength{\tabcolsep}{5pt}
\begin{tabular}{@{}lllrl@{}}
\toprule
Corpus & Task & Classes or largest groups & Test $n$ & Test distribution \\
\midrule
\multirow{3}{*}{MCV} & Sex/gender & male, female & 1,500 & 779 / 721 \\
 & Age & teens, twenties, thirties, forties, fifties & 1,237 & 263 / 262 / 262 / 263 / 187 \\
 & Native/accent & England, US, India/S. Asia, Canada, Australia & 1,500 & 404 / 403 / 403 / 162 / 128 \\
\addlinespace
\multirow{3}{*}{SAA} & Sex & male, female & 428 & 233 / 195 \\
 & Age & decade bins from 0--9 to 80--89 & 428 & largest: 20--29 (177), 30--39 (95) \\
 & Native language & English, Spanish, Arabic, Mandarin, others & 428 & largest: 117 / 36 / 22 / 15 \\
\bottomrule
\end{tabular}
\caption{Evaluation subsets. MCV age has 1,500 identifiers but 1,237 valid five-bin labels. SAA native language has many small classes; only the largest are listed.}
\label{tab:data}
\end{table*}

\subsection{Models and activation extraction}

Whisper-medium \footnote{\url{https://huggingface.co/openai/whisper-medium}} is an autoregressive encoder--decoder system over log-Mel features.
We pad or truncate its input to 3,000 frames and record encoder hidden states.
HuBERT-L \footnote{\url{https://huggingface.co/openai/whisper-medium}} and Wav2Vec2-L \footnote{\url{https://huggingface.co/facebook/wav2vec2-large-960h}} are CTC recognisers over 16~kHz waveforms; audio is capped or zero-padded to 30 seconds.
Each model exposes 25 hidden-state indices in the local Hugging Face interface: index 0 is the state before the first transformer block and indices 1--24 are block outputs.
All final hidden widths are 1,024, and no ASR parameters are trained.

For the main analysis, we average each hidden sequence over time to obtain one vector per clip and layer.
This makes layer-wise probing and direction estimation tractable across large subsets, but it discards temporal structure.
The saved extraction pipeline also averages padded positions without a mask, a limitation revisited below.

\subsection{Layer-wise probing}

For every model, task, and hidden-state index, we standardise features with a \texttt{StandardScaler} fitted on training data only and train one linear head.
The sex/gender probe is binary; age and native/accent probes initially retain all selected classes.
The MCV probes use Adam with learning rate $10^{-3}$, weight decay $10^{-8}$, weighted sampling, a 10\% validation split, and early stopping within 2,000 epochs.
SAA binary probes use learning rate $3\times10^{-4}$, weight decay $10^{-4}$, dropout 0.1, and longer patience.
All reported runs use seed 42.

We report macro-F1 because it gives each class equal weight even when test supports differ.
Uniform-random macro-F1 is computed using the observed supports and serves as a simple reference.
Probe performance is used to describe accessibility and, in some exploratory analyses, to compare layer rankings; it does not select every final steering layer.

\subsection{Baseline ASR and group gaps}

Each clip is first decoded with the unmodified recogniser.
References and hypotheses are lowercased, stripped of punctuation and symbols, and whitespace-normalised before WER.
Whisper uses 30-second chunks, six-second stride, and explicit English transcription generation.
HuBERT and Wav2Vec2 use their default CTC decoding paths without an external language model.

For each intervention family, two classes are selected: a source class with higher baseline WER and a lower-WER target class with sufficient support.
Examples include MCV teens$\rightarrow$forties and India/South-Asia$\rightarrow$England, and SAA Spanish$\rightarrow$English.
The comparison is not counterfactual: recording conditions, utterance difficulty, speaking rate, and other factors may contribute to the gap.

\subsection{Steering directions}

Let $h_i^{\ell}\in\mathbb{R}^{1024}$ be the mean-pooled training activation of clip $i$ at hidden-state index $\ell$, and let $S$ and $T$ denote source and target classes.
The centroid direction is:
\begin{equation}
v_{\mathrm{ctr}}^{\ell}=
\frac{1}{|T|}\sum_{i\in T}h_i^{\ell}-
\frac{1}{|S|}\sum_{i\in S}h_i^{\ell}.
\label{eq:centroid}
\end{equation}
For the probe direction, let $w_T-w_S$ be the weight difference of a binary probe trained on standardised features and let $\sigma$ contain the training feature standard deviations.
We map the direction back to activation coordinates as
\begin{equation}
v_{\mathrm{probe}}^{\ell}=(w_T-w_S)\oslash\sigma,
\label{eq:probe}
\end{equation}
where $\oslash$ is element-wise division.
This conversion accounts for the physical scale hidden by standardisation.

At inference time, a forward pre-hook adds the normalised direction to every time step at one encoder layer:
\begin{equation}
H^{\ell}\leftarrow H^{\ell}+\alpha\frac{v^{\ell}}{\|v^{\ell}\|_2}.
\label{eq:inject}
\end{equation}
Only clips with the known source metadata label receive the intervention; target clips are decoded in the same run but are not pushed.
This sample conditioning makes the experiment a mechanism diagnostic rather than a deployable correction.

\subsection{Selection, uncertainty, and trajectories}

Exploratory sweeps vary model, corpus, class pair, vector type, layer, strength, and direction.
Promising settings and explicit probe-best or adverse controls are then decoded again with matched baseline and steered outputs.
For source clips we report
\begin{equation}
\Delta_S=\mathrm{WER}_{\mathrm{steered},S}-\mathrm{WER}_{\mathrm{base},S},
\end{equation}
so negative values indicate fewer errors.
Each of the 22 targeted reruns uses 2,000 paired-bootstrap resamples and a 95\% percentile interval.
The descriptive $q_{\mathrm{boot}}$ is the fraction of resampled deltas that are non-negative.
It is neither a null-centred $p$-value nor a correction for the earlier search.

Trajectory diagnostics rerun the encoder with and without the hook and apply saved binary probes to each downstream layer.
For true source clips, we track mean target probability and the target rate, i.e., the fraction classified as the target class.
These quantities test whether a local intervention propagates in probe space; they do not show that a speaker attribute has literally changed.

\section{Results}

\subsection{Metadata is readable at different depths}

Table~\ref{tab:probe} reports the best MCV macro-F1 across 25 hidden-state indices.
Sex/gender is highly recoverable in all three encoders, with best F1 from 0.924 to 0.941.
Native/accent is also substantially above chance, especially in Whisper.
Five-class age is weaker but remains above the support-aware uniform-random baseline.

The best depth depends on both task and architecture.
Whisper peaks at indices 9, 13, and 17 for sex/gender, age, and native/accent; HuBERT peaks at 12, 8, and 13; Wav2Vec2 peaks at 10, 6, and 11.
There is therefore no single layer that contains all speaker-linked information most cleanly.
The age result should be read as recoverable age-correlated variation under coarse metadata bins, not as a robust age estimator.

\begin{table}[t]
\centering
\small
\setlength{\tabcolsep}{3.3pt}
\begin{tabular}{@{}lrrrr@{}}
\toprule
Task & Whisper & HuBERT & W2V2 & Chance \\
\midrule
Sex/gender & 9/.941 & 12/.939 & 10/.924 & .500 \\
Age & 13/.397 & 8/.375 & 6/.354 & .199 \\
Native/accent & 17/.696 & 13/.649 & 11/.544 & .190 \\
\bottomrule
\end{tabular}
\caption{Best MCV macro-F1. Cells give hidden-state index/F1. Chance is support-aware uniform-random macro-F1.}
\label{tab:probe}
\end{table}

\subsection{Baseline gaps vary by model and corpus}

Table~\ref{tab:gaps} shows the absolute WER difference for the binary pairs later considered for steering.
MCV sex/gender gaps are modest relative to the age and native/accent pairs, whereas SAA native-language gaps are large for all models.
For example, HuBERT produces 11.957\% WER for Spanish-source clips and 3.295\% for English clips, an 8.662-point gap.
The corresponding Wav2Vec2 gap is 11.000 points.

The same pair is not equally difficult for every model.
Whisper's SAA age gap is only 0.133 points and reverses the intended 30--39 to 20--29 direction at the displayed precision, while HuBERT and Wav2Vec2 show larger positive gaps.
This model dependence cautions against treating metadata alone as the cause of an error difference.

\begin{table}[t]
\centering
\small
\setlength{\tabcolsep}{3pt}
\begin{tabular}{@{}llrrr@{}}
\toprule
Corpus & Pair/task & Whisper & HuBERT & W2V2 \\
\midrule
MCV & teens/forties & 6.437 & 9.871 & 14.942 \\
MCV & India/England & 3.151 & 9.296 & 16.390 \\
MCV & male/female & .764 & 2.379 & 3.545 \\
SAA & 30--39/20--29 & .133 & .588 & 1.008 \\
SAA & Spanish/English & 2.741 & 8.662 & 11.000 \\
SAA & male/female & .612 & 1.442 & 2.447 \\
\bottomrule
\end{tabular}
\caption{Absolute baseline WER gaps (percentage points) for selected binary pairs.}
\label{tab:gaps}
\end{table}

\subsection{Selected WER effects are small and mixed}

\begin{figure*}[t]
    \centering
    \includegraphics[width=0.98\linewidth]{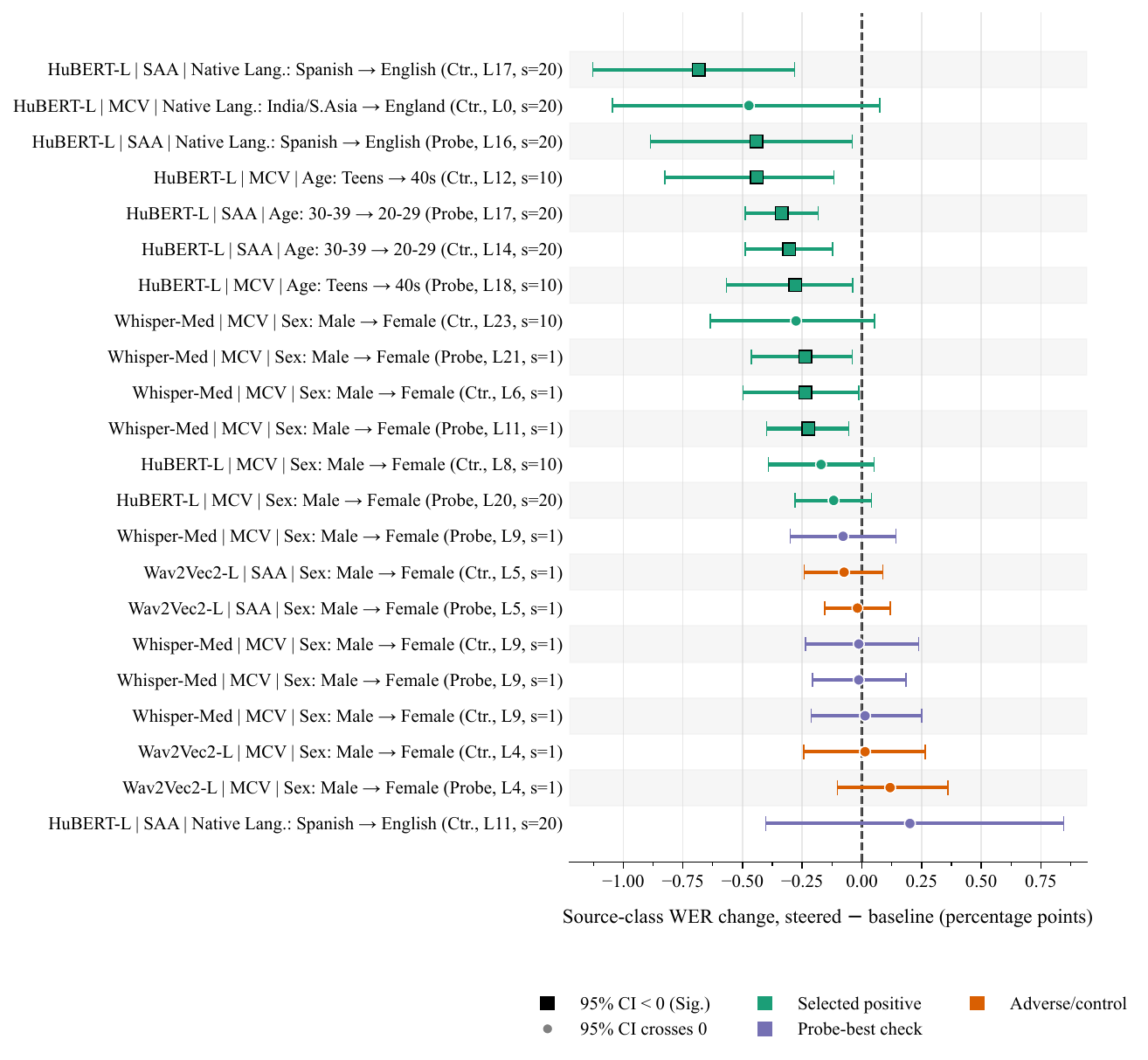}
    \caption{Post-selection source-class WER changes for all 22 targeted reruns (steered minus baseline, percentage points). Negative values favour steering; horizontal bars are paired-bootstrap 95\% intervals. Squares mark intervals wholly below zero and circles intervals crossing zero. Row labels give model, evaluation set, source$\rightarrow$target direction, vector type, injection layer, strength, and source sample size. Because configurations were chosen after exploratory sweeps, the intervals are descriptive follow-up evidence rather than family-wise confirmatory tests.}
\label{fig:forest}
\end{figure*}

Figure~\ref{fig:forest} reports all 22 targeted follow-up reruns, rather than only the supported cases.
Eighteen have negative source-class point estimates, but only nine have paired-bootstrap intervals entirely below zero.
Nine other negative estimates are inconclusive, and four estimates are non-negative.
Because the candidates were chosen after exploratory sweeps, these proportions are not unbiased estimates of how often steering works.

Every absolute reduction is below one point.
The largest is R6, HuBERT SAA Spanish$\rightarrow$English centroid steering: source WER changes from 11.957\% to 11.272\%, a saved delta of $-0.684$~\pp{} with $q_{\mathrm{boot}}=.0010$ and only 36 source clips.
R3, HuBERT MCV teens$\rightarrow$forties centroid steering, changes WER by $-0.440$ points on 263 clips.
Supported Whisper sex/gender effects are smaller, about $-0.22$ to $-0.24$ points.
The forest also exposes settings such as R16 and R17 whose point estimates are adverse.

The magnitude matters relative to the original gap.
R6 closes roughly 0.684 of an 8.662-point Spanish--English HuBERT gap.
R3 closes 0.440 of a 9.871-point teen--forties gap.
These are detectable changes in selected settings, not parity and not evidence that the metadata direction uniquely caused the benefit.

\subsection{Probe movement can disagree with WER}

\begin{figure*}[ht]
\centering
\begin{minipage}{0.48\textwidth}
  \centering
  \includegraphics[width=\linewidth]{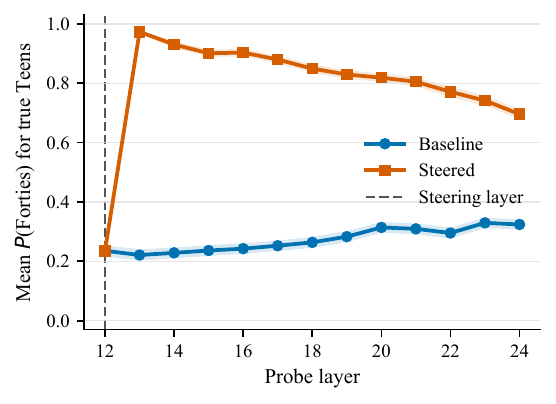}
\end{minipage}\hfill
\begin{minipage}{0.48\textwidth}
  \centering
  \includegraphics[width=\linewidth]{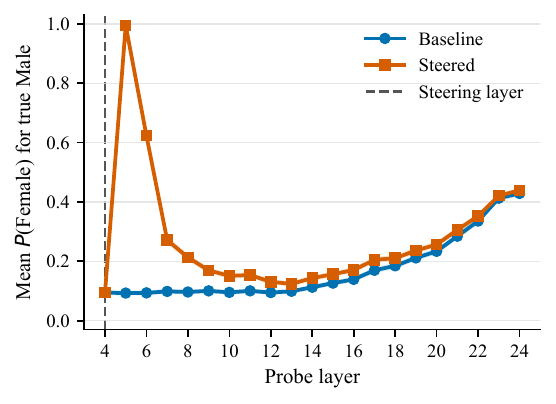}
\end{minipage}

\caption{Downstream probe trajectories expose the gap between representation movement and ASR improvement. Left: R3 HuBERT centroid steering from teens to forties at index 12 (strength 10; $n=263$ source clips) produces a large, persistent target-probability increase, while source WER improves by 0.440~\pp{}. Right: R17 Wav2Vec2 probe steering from male to female at index 4 (strength 1; $n=779$) produces an almost deterministic immediate shift that largely decays, while source WER worsens by 0.118 points. Lines are baseline and steered passes; dashed lines mark intervention indices.}
\label{fig:trajectories}
\end{figure*}

Table~\ref{tab:trajectory} and Figure~\ref{fig:trajectories} connect the internal and task-level measurements.
R2, a Whisper sex/gender direction transferred from SAA to MCV, raises the next-layer target rate from 5.52\% to 99.87\%.
The shift remains strong at the final encoder index (5.13\% baseline versus 96.53\% steered), but WER improves by only 0.237 points.

R3 shows a similarly large HuBERT age shift.
At index 13, mean target probability changes from 0.221 to 0.973 and target rate from 17.87\% to 98.48\%.
The probability remains elevated at the final index (0.324 to 0.696), yet the WER reduction is 0.440 points.
R6 moves fewer Spanish-source clips to the English probe class (2.78\% to 47.22\%) but has the largest WER reduction among the selected reruns.
Thus a larger internal shift does not imply a larger recognition gain.

R17 is the clearest failure.
Wav2Vec2 male$\rightarrow$female probe steering at index 4 raises the next-index target rate from 8.09\% to 99.87\%.
The effect then decays: at the final index, target rates are 39.67\% and 40.56\%.
Meanwhile source WER changes from 36.760\% to 36.879\%, a saved $+0.118$-point adverse estimate.
The intervention succeeds locally under the probe while failing as ASR control.

\begin{table*}[t]
\centering
\small
\setlength{\tabcolsep}{4pt}
\begin{tabular}{@{}lllclrrrrr@{}}
\toprule
Run & Model & Set/task & Vector & Source$\to$target & $n$ & Base & $\Delta$ & $q_{\rm boot}$ & Target rate \\
\midrule
R2 & Whisper & M/S sex & Probe & male$\to$female & 779 & 12.806 & $-.237$ & .0080 & 5.52$\to$99.87 \\
R3 & HuBERT & M/M age & Centroid & teens$\to$forties & 263 & 27.618 & $-.440$ & .0055 & 17.87$\to$98.48 \\
R6 & HuBERT & S/S native & Centroid & Spanish$\to$English & 36 & 11.957 & $-.684$ & .0010 & 2.78$\to$47.22 \\
R17 & W2V2 & M/M sex & Probe & male$\to$female & 779 & 36.760 & $+.118$ & .8646 & 8.09$\to$99.87 \\
\bottomrule
\end{tabular}
\caption{Contrasting trajectory reruns. ``Set'' gives evaluation/vector data (M=MCV, S=SAA). Base and $\Delta$ are source-class WER in percent and percentage points. Target rate is measured at the first diagnostic layer after injection.}
\label{tab:trajectory}
\end{table*}

\subsection{Probe quality is not a general layer selector}

If accessibility identified causal control points, layers with stronger binary probes should tend to yield larger steering benefits.
The result is configuration-dependent.
For Whisper MCV sex/gender centroid steering, the Spearman correlation between probe accuracy and best saved steering benefit across strengths is essentially zero ($\rho=.0041$, two-sided permutation $p=.9858$); the probe peaks at index 9 and steering at 23.
The probe-vector version is weakly negative ($\rho=-.2908$, $p=.1592$).

HuBERT MCV age is a counterexample where rankings align: centroid steering gives $\rho=.6430$, $p=.0014$, with binary-probe-best index 13 and steering-best index 12.
SAA HuBERT age probe steering is also positive ($\rho=.5072$, $p=.0110$).
These cases show that probes can narrow exploration in some settings, but no architecture-independent layer-selection rule emerges.

\subsection{Utterance-level outcomes are mixed}

Aggregate WER hides which words change.
Three paired examples illustrate the range without serving as evidence on their own.
In R21, the baseline transcribes the place name ``Padstow'' as ``pad style,'' while the steered hypothesis matches the reference.
In R1, both outputs correctly transcribe the same utterance.
In adverse R17, the baseline is correct but steering changes ``Berlin work became practically'' to ``Berlinwork became practicalaly.''
Table~\ref{tab:qualitative} shortens the excerpts for space.

\begin{table*}[t]
\centering
\small
\setlength{\tabcolsep}{3pt}
\begin{tabular}{@{}p{0.04\textwidth}p{0.09\textwidth}p{0.25\textwidth}p{0.25\textwidth}p{0.25\textwidth}@{}}
\toprule
Run & Case & Reference cue & Baseline cue & Steered cue \\
\midrule
R21 & Improved & opposite Padstow & opposite pad style & opposite Padstow \\
R1 & Unchanged & the boy swore that every time & correct & correct \\
R17 & Harmful & Berlin work became practically & correct & Berlinwork became practicalaly \\
\bottomrule
\end{tabular}
\caption{Illustrative paired transcription excerpts. Examples reveal error types but do not establish an aggregate effect.}
\label{tab:qualitative}
\end{table*}

The mixed examples are consistent with the rerun statistics.
A constant speaker-linked direction can repair one local error, leave many clips unchanged, and introduce errors elsewhere.
Paired resampling aggregates these heterogeneous outcomes, while trajectory probes reveal whether the internal perturbation persists.
Neither replaces a user-centred assessment of error severity.

\section{Discussion}

\subsection{Three validity tests for mitigation}

The experiments separate three questions that representation-based fairness interventions should answer.
First, is the attribute linearly readable?
Second, does an intervention move later states in the intended representational direction?
Third, does task behaviour improve for the affected group without unacceptable harm elsewhere?
Our evidence gives a strong conditional ``yes'' to the first, mixed answers to the second, and weak, post-selected evidence for the third.

This separation changes how a steering result should be interpreted.
A target-like probe flip is a mechanism check, not a fairness metric.
It can reflect movement along correlated acoustic dimensions, removal of content useful for decoding, or a transient displacement later blocks undo.
The differing Whisper, HuBERT, and Wav2Vec2 trajectories also suggest that encoder geometry cannot be evaluated independently of the downstream decoder.
Whisper's autoregressive decoder may absorb or ignore an encoder perturbation differently from CTC decoding, while later self-supervised blocks can attenuate a locally large shift.

\subsection{Implications for fairness claims}

The source classes were chosen because their observed WER was higher, but the comparison classes are not causal counterfactuals.
The direction male$\rightarrow$female or Spanish$\rightarrow$English is analytical shorthand, not a claim that one group ought to sound like another.
Indeed, treating lower-error groups as representational targets can reproduce assimilationist assumptions if divorced from this diagnostic context.

The true metadata label also determines which clips receive steering.
That makes the experiment useful for controlled analysis but unsuitable as a deployment recipe.
A deployed recogniser would not normally know a speaker's age, sex/gender, or native-language label before transcription; collecting or inferring it creates privacy and fairness risks of its own.
The small improvements here do not justify such a requirement.

\subsection{What stronger evidence would require}

Simple directions remain valuable because they are transparent and auditable baselines.
Future work should nevertheless separate discovery from evaluation: estimate directions on training data, select layers and strengths on a validation set, and reserve a final test set for fairness outcomes.
A systematic placebo suite should include norm-matched random directions, shuffled-label centroids, opposite directions, and within-class split directions.
Applying each direction to target and unrelated groups would expose collateral harm.

Evaluation should also go beyond WER.
Metrics such as match error rate (MER) and word information lost (WIL) \citep{morris04_interspeech}, alongside phoneme error, semantic changes, named-entity errors, calibration, and user-relevant severity, can distinguish an inconsequential substitution from an accessibility-critical failure.
Temporally local interventions may better match accent or age cues than adding one vector to every frame.
For Whisper, decoder cross-attention is another plausible intervention point.
The core criterion should remain behavioural: an interpretable or steerable feature is useful only if it supports robust improvements under held-out evaluation and appropriate controls.

\section{Conclusion}

Speaker-linked metadata is linearly accessible across three major ASR encoders, but accessibility does not make the same directions reliable control mechanisms.
Selected steering configurations sometimes yield small WER reductions and large internal probe shifts, yet the two effects frequently diverge.
The strongest local probe shift in our diagnostic set can accompany worse recognition.
Bias-mitigation studies should therefore evaluate readability, propagation, and downstream utility separately and treat success at any one level as insufficient.

\section*{Limitations}

The metadata labels are imperfect and reductive.
MCV fields are self-reported and incomplete; SAA classes are small; binary male/female labels exclude non-binary identities; and native-language, accent, age, and sex/gender are not interchangeable with acoustic properties.
Our labels should be interpreted only as dataset-derived groupings.

The intervention is a diagnostic rather than deployable mitigation.
It uses the true source label to decide which clips to steer, whereas a real recogniser would not generally know that label before transcription.
Using inferred or declared sensitive attributes at deployment would create additional fairness and privacy risks.

The selected reruns follow exploratory sweeps over models, datasets, pairs, directions, layers, and strengths.
Paired bootstrapping quantifies uncertainty for a chosen configuration but does not remove selection bias or control a search-wide error rate.
A separate validation set and untouched final test set are needed for confirmatory claims.
The final suite also lacks systematic placebo directions and a complete analysis of effects on non-source groups.

The SAA split audit is weaker than MCV's explicit \texttt{client\_id} check, and several SAA comparisons have small source groups, including 36 Spanish clips.
The group comparisons are not counterfactual; recording quality, sentence content, speaking rate, and other difficulty factors may contribute to WER differences.

Mean pooling removes temporal structure, and the saved pipeline averages padded hidden sequences without masking.
Adding one constant vector at every time step may therefore exaggerate a global speaker direction and poorly match phoneme-local accent or age cues.
The evaluation covers two English datasets and three pretrained systems, not multilingual speech LLMs or production conditions.
Finally, WER treats all word errors alike and does not measure semantic severity or user impact.

\section*{Ethical Considerations}

This study analyses labels linked to socially sensitive attributes.
High probe scores must not be interpreted as accurate inference of identity, nor as justification for demographic profiling.
The source-to-target notation follows an observed WER contrast and does not imply that a higher-error group should acoustically conform to a lower-error group.
Any future deployment-oriented mitigation should avoid requiring sensitive labels where possible, obtain appropriate consent, assess intersecting groups, and test whether gains for one group impose harms on others.
The small, post-selected effects reported here are explicitly presented as a warning against premature fairness claims.

\bibliography{custom}

\end{document}